# Modeling Normal Is All You Need: Joint Latent Clustering for Anomaly Detection in Multimodal Cyber-Physical Systems

**Alexander Apartsin [1] and Yehudit Aperstein [2,*]**

[1] School of Computer Science, Faculty of Sciences, Holon Institute of Technology (HIT), 52 Golomb St., Holon 5810201, Israel; alexanderap@hit.ac.il
[2] Intelligent Systems, Afeka Academic College of Engineering, 218 Bnei Efraim St., Tel Aviv 6910717, Israel; apersteiny@afeka.ac.il
[*] Correspondence: apersteiny@afeka.ac.il

**Abstract**

A cyber-physical system (CPS) can enter a faulty state that is individually normal on every sensor and reconstructs accurately, yet is improbable under normal joint operation. This exposes the central weakness of reconstruction-based detection: reconstruction measures whether a state can be reproduced (reachability), not whether normal operation is likely to occupy it (probability). We model CPS normal behavior as a union of many imbalanced operating regimes, ten assumptions we call *Massive, Implicit, Imbalanced Multimodality* (MIIM). Our detector, LatAD, jointly learns a latent and a Gaussian-mixture clustering of these regimes (VaDE) and scores anomalies by density rather than reconstruction. Because a CPS is an assembly of coupled subsystems, we factorize that density over correlation-community subsystems and combine per-community surprises by a cohesion-weighted, sparsity-adaptive statistic, concentrating a local fault a global density dilutes. Evaluated with raw point-wise metrics and a difficulty split isolating the stealthy faults a per-channel threshold misses, LatAD attains the best AUROC on three real CPS benchmarks (WADI 0.862, HAI 0.949, SWaT 0.993) and leads the difficult subset of all three, notably on HAI (0.849; a significant +0.09 over the next-best baseline, 95% CI [0.046, 0.160]); the reconstruction-based USAD and TranAD fall to 0.30–0.48 on these reconstructable-but-improbable faults.

## 1 Introduction

A cyber-physical system can be faulty even when every sensor reads within its normal range and a deep model reconstructs the full sensor vector accurately. The reason is that reconstruction and normality answer different questions: reconstruction asks whether a state can be reproduced, whereas detection needs to know whether normal operation is ever likely to occupy it. In a tightly coupled plant, vehicle, or machine, normal behavior is spread across many operating regimes shaped by physics, actuator limits, and control logic, so a correlation break, or a state that falls between two legitimate regimes, can be perfectly reconstructable yet deeply improbable. These are the faults that matter most for condition monitoring and predictive maintenance: early warnings that stay inside familiar sensor ranges while violating the joint structure of normal operation. This paper detects them by modeling that joint structure, scoring probability rather than reconstruction, and, because faults are as local as the physics that produces them, factorizing that probability over the plant's own subsystems.

A modern CPS (a water-treatment plant, an industrial control loop, a rotating machine) couples hundreds of networked sensors, controllers, and actuators into a unit that continuously senses its own state and acts on it: the sensing-and-actuation fabric of the industrial Internet of Things (IoT). This complexity puts its faults beyond enumeration: no test campaign can sample every combination of component, control regime, and operating condition in which a fault can arise, and many faults develop slowly, as wear or drift with no obvious symptom, until they surface as unplanned downtime or a safety-critical failure. Because the faults cannot be specified in advance, they must be discovered from the system's own behavior as it operates, the task of anomaly detection [1]. That behavior is richly observable: vibration, current, temperature, pressure, and position, together with internal control signals, stream continuously as IoT telemetry and carry an early signature of almost any developing fault.

The task is often framed as characterizing the anomalies, but that framing is misleading. With a representative set of labeled faults the task would reduce to supervised classification; in practice genuine faults are rare and the few on record seldom span the ways a system can fail [2], so they cannot define the decision boundary, and they cannot even be spent reliably on model selection. The usable information lies on the other side of the boundary, in the normal data, which is abundant and rarely exploited in full. Modeling that normal law faithfully is therefore the hard part of the problem, and cyber-physical normal is a law of a particular kind: shaped by deterministic physics, hard actuator and setpoint limits, and engineered control, it is the union of many bounded, curved operating regimes, some common and many rare, each with legitimately low-density fringes, rather than a single statistical blob. A detector that mis-models this structure fails however it thresholds.

A fault is anything that leaves this union of normal regimes, and the operationally difficult ones stay *inside* the marginal envelope of every channel while breaking the joint structure: a correlation break, a between-regime pocket, a history-inconsistent state. These faults are invisible to any per-channel range rule, and, as we show, to a reconstruction residual as well, because the question is not whether a flexible decoder can reproduce the state but whether normal operation is likely to occupy it. Reconstruction measures reachability; detection needs probability.

Two obstacles have obscured this failure mode. The first is modeling. If normal is a union of many imbalanced regimes, a global anomaly model under-fits the regime structure, a single deep autoencoder alarms on rare-but-valid regimes (the trust-eroding false positive), and a reconstruction residual is blind to exactly the joint-structure faults that matter, because a flexible decoder reconstructs them faithfully.

The second is evaluation. The dominant metric in the literature, point-adjusted F1, inflates scores so severely that a *random* anomaly score can beat every published deep model [3, 4]; and even under raw metrics, public CPS benchmarks are dominated by anomalies that a one-line univariate rule already separates [5], so a high headline number certifies little. Progress on the faults that matter cannot be measured until the easy majority is separated from the discriminative minority.

This paper makes three contributions. **(1)** We state the structural assumptions of CPS normal data as an explicit list, A1–A10, abbreviated MIIM (*Massive, Implicit, Imbalanced Multimodality*), and tie each assumption to a concrete choice for modeling normal (§3). **(2)** We build **LatAD**, a detector whose whole design is modeling that normal law: a jointly learned latent with explicit Gaussian-mixture regime clustering (VaDE), scored in the latent by a mixture-density head and a rare-regime-safe nearest-component likelihood (§4). Its key mechanism is demoting reconstruction from the default score, since a flexible decoder reconstructs the hard faults faithfully; reconstruction re-enters only through an optional residual head, switched on by a gate that tests generalization on held-out normal data. Its headline realization factorizes this density over the plant's own subsystems, recovered unsupervised as communities of the normal-data correlation graph, and combines the per-subsystem surprises by cohesion-weighted Higher Criticism, so a local fault that a single global density dilutes is concentrated instead (§4.4). **(3)** We evaluate under a fair protocol (raw point-wise metrics, difficulty stratification, train-normal-only calibration) on three real CPS attack benchmarks, comparing against classical baselines and the deep detectors USAD and TranAD re-computed with the same raw metrics (§5–6). LatAD attains the best overall AUROC on every dataset and leads the difficult subset of all three, most decisively on HAI, where the difficult faults break joint structure, the deep detectors collapse to near chance, and LatAD's margin is statistically significant.

Because aggregate benchmark scores are dominated by easy marginal excursions, the central test throughout is performance on the difficult subset, the anomalies a simple per-channel detector cannot separate, not the headline number. The study covers three datasets with a single trained configuration per dataset; AUROC is reported as a five-seed mean, and the best-F1 threshold is a test-swept oracle disclosed in §5.

## 2 Related work

Anomaly detection is the unsupervised or semi-supervised task of flagging departures from a model of normal behavior, adopted precisely because labeled anomalies are too scarce to train a classifier [1, 6, 2]. On time series it must additionally respect temporal and cross-channel dependence [7], and on cyber-physical and industrial control systems it underpins physics-aware intrusion and fault detection [8]. LatAD sits squarely in this line: unsupervised, trained on normal only, and specialised to the multimodal structure of cyber-physical normal behavior.

### 2.1 *The illusion of progress in time-series anomaly-detection evaluation*

A line of critical work shows that the standard evaluation of time-series anomaly detection is unreliable. Kim et al. [3] demonstrate that the ubiquitous *point-adjustment* protocol (marking an entire ground-truth anomaly segment as detected if any single point in it is flagged) inflates F1 so drastically that a *random* anomaly score achieves higher point-adjusted F1 than every state-of-the-art model on SWaT and WADI. Garg et al. [4] independently reproduce this inflation and show that simple baselines (PCA, channel-wise autoencoders) beat elaborate deep models once metrics are made rigorous, and that deep methods often fail to detect even simple anomalies. Doshi et al. [9] diagnose the point-adjustment reward structure and propose a rectified evaluation, and proximity- and event-aware metrics such as PATE [10] move beyond point adjustment altogether. The critique now spans metrics and benchmarks alike. Wu and Keogh [5] argue that many popular benchmarks are flawed (triviality, mislabelling and unrealistic anomaly density) creating an illusion of progress; and a recent benchmark analysis [11] finds that anomaly segments in multivariate CPS benchmarks are mostly univariate (SWaT and WADI sit among the datasets whose anomalies deviate univariately on close to all timesteps, with no long cross-channel-only segments), so a flat univariate model matches channel-dependent deep detectors. Recent large-scale benchmarks confirm that detector rankings are strongly dataset-dependent [12], which is why we read every method per dataset and per difficulty subset rather than by a single headline number.

A parallel critique targets the dominant scoring mechanism itself. Reconstruction-based detectors assume that a model trained to reconstruct normal data reconstructs anomalies worse; Bouman and Heskes [13] show this assumption is unreliable, because anomalies lying far from normal data can be reconstructed perfectly in practice, and Gong et al. [14] earlier showed that a deep autoencoder generalizes well enough to rebuild anomalies and miss them. This motivates scoring in a clustered latent rather than through a reconstruction residual. A position paper by Sarfraz et al. [15] draws the two threads together, characterizing the field as plagued by flawed evaluation metrics, inconsistent benchmarking, and unjustified design choices, and showing that state-of-the-art deep detectors effectively learn linear mappings that simple baselines reproduce; Wagner et al. [16] re-audit the widely used multivariate benchmarks, discard the ones with erroneous labels, and note the absence of any standard evaluation metric.

LatAD adopts this critique as its evaluation protocol: raw point-wise AUROC and F1 with no point adjustment for any method; a trivial-detector difficulty split that separates the discriminative difficult subset from the easy majority that carries headline numbers; and re-computation of the deep SOTA baselines (USAD and TranAD) under the *same* raw metrics rather than their point-adjusted published figures (§5).

### *2.2 Latent and clustering models for anomaly detection*

Our representation is Variational Deep Embedding (VaDE) [17], a VAE with a Gaussian-mixture latent prior that learns representation and clusters jointly. DAGMM [18] similarly couples an autoencoder with a Gaussian-mixture energy for anomaly detection; we borrow its variance-collapse regularisation but score against explicit per-regime Gaussians in a jointly learned latent rather than a learned-Gaussian energy (§4). Our whitened residual uses the Ledoit–Wolf shrinkage covariance estimator [19]. Classical baselines are Isolation Forest [20] and a deep autoencoder [21], both fitted on the same window features, together with a trivial per-channel range detector that also defines the easy/difficult split.

### *2.3 Deep multivariate CPS anomaly detection*

USAD [22] couples two adversarially trained autoencoders; TranAD [23] uses deep transformer networks with adversarial and self-conditioning training; GDN [24] learns a graph over channels and detects deviations from learned inter-channel relations. These define a widely used deep SOTA on SWaT/WADI-style benchmarks. More recent detectors refine rather than abandon reconstruction: SensitiveHUE [25] models per-variable heteroscedastic uncertainty to sharpen sensitivity to normal patterns, and CATCH [26] learns channel correlations through frequency-domain patching. Both remain reconstruction-based, so the mechanism analysis of §7 suggests they inherit the same blind spot on the reconstructable-but-improbable joint faults that define our difficult subset. We benchmark USAD and TranAD, the reconstruction- and transformer-based deep detectors, under the same raw, difficulty-stratified metrics over five seeds (the re-scored comparison is in §6), and additionally re-run the cross-channel GDN under the identical raw protocol where its per-window graph training was tractable (SWaT). A raw-metric evaluation of the newer reconstruction detectors (SensitiveHUE, CATCH) is left to future work.

Beyond public benchmarks, sensor-driven condition monitoring and fault prediction in operational CPS motivate modeling the multivariate sensor stream directly: vibration-based machine diagnostics [27, 28], structural-health monitoring of rotorcraft components [29], and fault prediction in rotorcraft flight controls [30]. LatAD generalizes this applied line to unsupervised, multimodal-latent anomaly detection.

### 2.4 Benchmark datasets

We evaluate on three real CPS attack testbeds. WADI [31] is a 123-channel water-distribution testbed from iTrust, and SWaT [32] is its 51-channel water-treatment sibling; the two are the de facto standard multivariate CPS benchmarks and the primary targets of the evaluation critique above [3, 5]. HAI [33] is a 59-channel hardware-in-the-loop industrial control testbed spanning several coupled processes. All three carry real executed attacks and are strongly multimodal (BIC-optimal regime counts $K^* \geq 22$; §3), so the difficulty-stratified results read directly against the MIIM structure of each dataset.

## 3 The MIIM assumptions about CPS normal data

We model the normal law of a CPS as a mixture over reachable operating regimes, $p(x) = \sum_{k=1}^{K} \pi_k \, p_k(x)$ where each component $p_k$ is a bounded, oriented, curved patch of observation space. The ten assumptions of Table 1 describe the qualitative structure of this mixture; each is grounded in a physical property of CPS operation and each motivates a specific design choice in the detector of §4. We call the resulting property *MIIM*: the sample is **M**assive, the regime structure is **I**mplicit (unlabeled and only approximately recoverable), the regime occupancy is **I**mbalanced, and normal is **M**ultimodal. Two consequences matter for detection. Normality is multimodal in regime space; and because the few levers of A5 drive typed, correlated channel groups (A8), deviations are often local to one tightly coupled subsystem. LatAD therefore models regime probability at subsystem scale, not only at whole-plant scale (§4.4).

**Table 1.** The ten CPS normal-data assumptions, their physical origin, and the detector design choice each motivates. A1–A8 shape the instantaneous window; A9–A10 shape the trajectory.

| ID | Assumption | CPS rationale | Motivates (design choice) |
|---|---|---|---|
| A1 | **Regime mixture** (multimodality) | A plant or vehicle cycles through discrete operating regimes (idle, cruise, load steps); normal is their union, not one blob. | Model normal as an explicit mixture, not a global density: the latent GMM prior of VaDE. |
| A2 | **Regime explosion** (exponential regime growth) | A CPS is built from many interacting subsystems; each added element multiplies reachable regimes, so regime count grows with complexity. | Use a high number of components $K$ (and a high-resolution density head) rather than a handful. |
| A3 | **Hard envelopes** (bounded regimes; thin between-regime pockets) | Controllers hold each regime within setpoint/actuator bands; the space between regimes is passed through, rarely dwelt in, leaving thin anomalous pockets. | A between-regime point must score anomalous even when every channel is in range: the basin-agreement head (iv). |
| A4 | **Thin fringes** (intra-regime sparsity) | Within a regime the system dwells near a typical operating point and reaches the edges only occasionally; each regime holds legitimate low-density regions near its boundary. | Density inside a regime is non-Gaussian → a parametric high-resolution mixture density head, not one Gaussian per regime. |
| A5 | **Few levers** (low intrinsic dimension) | The independent control variables are far fewer than the observed channels; the rest follow deterministically or probabilistically. | A low-dimensional learned latent (dim 10–16 across datasets) suffices; faults appear as departures within that latent. |
| A6 | **Heavy tail** (Zipf imbalance) | Machines spend most of the record in a few steady regimes; startup, shutdown and rare manoeuvres occupy little of it, so an infrequent-but-normal point is easily mistaken for a fault. | Score against the *nearest* component, never the π-weighted mixture, so a rare-but-valid regime is not penalized (ii). |
| A7 | **Hidden regimes** (latent, implicit regimes) | True regimes are an emergent artifact of coupling and are not logged; regimes must be | Discover regimes jointly with the representation (VaDE), |

| ID | Assumption | CPS rationale | Motivates (design choice) |
|---|---|---|---|
| | | discovered approximately from raw streams. | unsupervised; never use regime labels. |
| **A8** | **Mixed signals** (heterogeneous, typed channels) | The bus carries sensors, actuators, discrete states and setpoints, each with its own range, resolution and correlated noise. | Per-feature standardization on train-normal; a whitened (Mahalanobis) residual that respects per-channel scale and cross-channel correlation (iii). |
| **A9** | **Many clocks** (multiscale dynamics) | Physical inertia sets per-channel time constants (thermal slow, electrical fast); only some regime→regime transitions are legal, each with a characteristic dwell. | Fixed-length windows with per-channel statistics as features; motivates explicit temporal features, left to future work (§7). |
| **A10** | **Path dependence** (history-conditioned normality) | A fault is defined relative to operational history: the same instantaneous state can be normal or faulty depending on how the system arrived there. | In principle a trajectory branch; on these snapshot-detectable public benchmarks (each fault visible in a single window, without history) it is inert (see §7), so the reported model is window-only. |

Two standing qualifiers apply throughout: the sample is **massive** ($N$ large) and the mixture is approximately **stationary** within a run. Measured on train-normal, the MIIM structure is present but *implicit* (A7): cluster sizes are heavy-tailed, and Bayesian-information-criterion (BIC) regime counts keep improving past two dozen components (BIC-optimal $K^*$ of 22, 24, and 25 on WADI, HAI, and SWaT). The regimes overlap on WADI and HAI (silhouette 0.06 and 0.08, on a scale where 1 is well-separated clusters and 0 is overlapping) but separate more crisply on SWaT (0.29), and this ordering tracks where the design pays off: the two overlapping datasets are exactly where the regime-community factorization gains most, since a single global density blurs local structure there, while crisp SWaT is a ceiling for every method (§6). This is the regime in which a jointly learned latent, scored by a mixture density and factorized over subsystems, pays off and a global detector or a raw reconstruction residual does not.

## 4 Method

We call the detector **LatAD**, and preview its pipeline before the details: windowed telemetry → correlation-community subsystems → one regime-density model per community plus a global expert → calibrated per-community tail probabilities → cohesion-weighted Higher Criticism → anomaly score. Two stages implement it. A representation stage, a VaDE, jointly learns a latent and a Gaussian-mixture regime prior; a scoring stage computes a stack of density heads in that latent, all calibrated on train-normal. The headline **LatAD (regime-community)** estimates this density per subsystem (§4.4); the base ablation **LatAD (global density)** estimates it once over the whole channel set. Because every community uses the same representation and heads, we describe those first, then the regime-community realization.

The reported detector realizes assumptions A1, A2, A4–A8 and architecturally supports A3. The mixture representation (A1, A7), the high-resolution density head (A2, A4), the low-dimensional latent (A5), the nearest-component score (A6) and per-feature standardization (A8) are always active; the whitened-residual head (A8) and the basin-agreement head (A3) are *auto-gated* on train-normal signals and switch on only where a dataset calls for them: the residual activates on HAI and SWaT, and the basin head activates only when train-normal regime assignments are ambiguous, which none of these three datasets trigger, so A3 is architecturally supported but inactive here. The gated heads did not reduce performance on any of these datasets, and the reported gains come from the always-active heads, with the auto-gated residual head adding to them on HAI and SWaT (Table 5). Assumptions A9–A10 shape the trajectory rather than the window and are not exercised by these snapshot-detectable public benchmarks; extending the detector to explicit multiscale-temporal features (A9) and history-conditioned, path-dependent scoring (A10) is left to future work (§7).

**One realization of the assumptions.** A1–A10 specify *what* a model of CPS normal must capture, not *how* to implement it; the detector below is one concrete realization of A1–A8, reported in full so that every choice is auditable, and §7 surveys the design space the same assumptions admit.

### *4.1 Representation: Variational Deep Embedding (VaDE)*

VaDE [17] is a variational autoencoder whose latent prior is a Gaussian mixture, so the representation and the operating-regime clusters are learned *together* (A1, A7) rather than one after the other. An encoder maps a window feature vector $x$ to a latent Gaussian $q(z \mid x) = N(\mu, \mathrm{diag}\,\sigma^2)$; a decoder reconstructs $\hat{x}$; and the prior is

$$p(z) = \sum_{c=1}^{K} \pi_c \; N(z \mid \mu_c, \mathrm{diag}\,\sigma_c^2),$$

with mixture parameters $\{\pi_c, \mu_c, \sigma_c\}$ learned jointly with the networks. The training objective is the negative evidence lower bound; per window it is

$$\mathcal{L} = \sum_j (x_j - \hat{x}_j)^2 + \beta \sum_c \gamma_c \left[ \mathrm{KL}(q \,\|\, N(\mu_c, \sigma_c^2)) + \log \frac{\gamma_c}{\pi_c} \right],$$

where $\gamma_c \propto \pi_c\, N(z \mid \mu_c, \sigma_c^2)$ are the responsibilities (the posterior probability that a window belongs to regime $c$). Three standard measures prevent the well-known cluster collapse in which many components are abandoned during joint training: a KL/cluster *warm-up* that anneals $\beta$ from 0 to 1 so the encoder settles on the pretrained mixture initialisation before the prior is pulled around; a DAGMM-style [18] penalty on tiny component variances (a variance floor plus a $\sum_c \sigma_c^{-2}$ term); and a slower learning rate on the mixture parameters so the GMM initialisation (fitted on the pretrained latent) is refined, not destroyed. Training proceeds as plain-VAE pretraining, GMM initialisation on the encoded means, then joint optimization of the full objective.

### 4.2 *Scoring: why we demote reconstruction*

The natural VaDE anomaly score is the joint negative log-likelihood: a whitened reconstruction residual plus a latent term. On the *difficult* faults (correlation breaks and between-regime pockets) this is exactly wrong. Table 2 decomposes the two terms by their difficult-subset AUROC on a single trained model per dataset.

**Table 2.** Difficult-subset AUROC of the two natural score terms (single trained model per dataset). The reconstruction residual is near chance on WADI (0.490) and, although informative on HAI (0.689), it stays below the latent term (0.760) and drags the summed joint score down; the latent likelihood carries the signal but a single diagonal Gaussian per regime is coarse. A multi-seed, three-dataset head ablation is reported in Table 5.

| Score term | WADI | HAI |
|---|---|---|
| (a) reconstruction residual $\frac{1}{2} r^{\top} \Sigma^{-1} r$ | 0.490 | 0.689 |
| (b) latent NLL (nearest diagonal component) | 0.663 | 0.760 |
| (a)+(b) joint NLL (naive VaDE score) | 0.494 | 0.689 |

The mechanism is that a flexible decoder *reconstructs the fault faithfully*: a correlation-break window has in-range marginals, so the decoder rebuilds it and the residual is near chance or reversed. Demoting reconstruction from the default score improves the difficult subset on the two joint-structure benchmarks (WADI and HAI). The base scoring stack therefore works entirely in the jointly learned latent; reconstruction re-enters only through the optional, auto-gated residual head (§4.3 iii), which fires only where it provably generalizes to held-out normal (HAI and SWaT here, not WADI).

### 4.3 *The scoring stack*

Let $z = \mu(x)$ be the encoder mean. The base score sums two latent heads, each z-normalized against its train-normal mean and standard deviation, so the operating scale is set by normal, not by the test batch.

(i) Latent mixture-density head (parametric KDE for non-Gaussian pockets)

Because each regime has a non-uniform interior with thin fringes (A4), one Gaussian per regime is too coarse. We fit a high-resolution diagonal Gaussian mixture on the train-normal latent ($M = 80$ components by default, distinct from the VaDE regime count $K$) as a parametric kernel-density estimate, and score a window by its density negative log-likelihood:

$$s_{\mathrm{dens}}(x) = -\log \sum_{m=1}^{80} w_m \; N(z \mid a_m, \mathrm{diag}\, b_m^2).$$

(ii) Nearest-component NLL (rare-regime-safe)

Under heavy-tailed imbalance (A6) a valid point in a rare, low-$\pi$ regime must not be flagged solely for being rare. We therefore use the *closest* component, not the $\pi$-weighted mixture:

$$s_{\text{near}}(x) = -\max_c \log N(z \mid \mu_c, \text{diag}\, \sigma_c^2).$$

Using the maximum over components rather than the Bayesian mixture sum keeps a rare-regime signal that the mixture sum washes out (A6): the nearest-component distance retains a deviation confined to a single sparsely populated regime, whereas the mixture sum averages it away. The base score is the sum of the two z-normalized heads, $s_0 = \text{z}(s_{\text{dens}}) + \text{z}(s_{\text{near}})$.

(iii) Optional responsibility-weighted whitened residual, auto-gated

On some datasets the fault *does* surface in reconstruction. The optional residual head reduces the per-channel residual $r$ by principal-component analysis (fit on train-normal) to $30$ dimensions, fits a Ledoit–Wolf-shrunk [19] precision $\Sigma_k$ per regime, and combines the per-regime whitened energies of that reduced residual by responsibility:

$$s_{\text{resid}}(x) = \sum_{k=1}^{K} \gamma_k \; r_k^{\top} \Sigma_k^{-1} r_k.$$

Here $r_k$ is the reduced residual whitened by regime $k$'s precision $\Sigma_k$, and $\gamma_k$ is the window's responsibility for regime $k$. The head is *auto-gated* by a held-out-normal generalization test: per-regime precisions are fitted on the first 80% of train-normal (split A) and scored on the last 20% (split B); if held-out normal scores much higher (the residual overfits or drifts), the ratio $q_{95}(B)/q_{95}(A)$ exceeds 1.5 and the head is switched off. Empirically the ratio is 5.24 on WADI (off), 1.17 on HAI (on), and 0.85 on SWaT (on): the residual carries the fault on HAI and SWaT but is dead on WADI, and the gate recovers this pattern from train-normal alone.

(iv) Optional basin-agreement rescue, auto-scaled

Between-regime pockets (A3) are the dangerous false negative on datasets whose regimes overlap. We perturb the latent with Gaussian noise $R$ times and measure *agreement*, the fraction of perturbed copies that keep the clean argmax regime. A rare-but-valid point sits deep in one basin (high agreement, demote); a between-regime point flips regimes under perturbation (low agreement, keep). The rescue subtracts a scaled, train-normal-calibrated agreement:

$$s(x) = s_0 + s_{\text{resid}} - \lambda\, \text{z}(\text{agree}(x)), \quad \lambda = \lambda_0 \cdot \max(0, \rho - \delta),$$

where $\rho$ is the fraction of train-normal windows that are ambiguous (maximum responsibility below 0.5) and $\delta$ a dead-zone. Although WADI and HAI regimes overlap in silhouette terms (§3), their per-window assignments stay confident: the ambiguous fraction $\rho \approx 0.05$ on all three datasets, so $\lambda = 0$ and the rescue is inactive. The basin head is a provision for heavily overlapping regimes and awaits data that exercises it. The reported model thus configures its heads purely from train-normal statistics, with no test-set tuning: the residual head fires on HAI and SWaT, and both optional heads stay off on WADI.

**Auto-gating summary.** Base heads (i–ii) run on every dataset. The residual head (iii) fires when it generalizes to held-out normal (HAI and SWaT: on; WADI: off). The basin head (iv) activates when a non-trivial fraction of train-normal windows have ambiguous regime assignments; that fraction is near zero on all three datasets here, so it is a no-op. All calibration uses train-normal only.

### *4.4 Regime-community density (the headline realization)*

The stack of §4.1–4.3 estimates one density over the whole channel set. Its headline realization estimates density *per subsystem* instead, and the reason is physical. A CPS is an assembly of coupled subsystems, process units, control loops, and conservation-linked flows, each binding a small group of sensors and actuators through tight local correlation while remaining only loosely coupled to the rest of the plant (WADI's parallel treatment trains, SWaT's six sequential stages, HAI's coupled boiler, turbine, and water processes). Mass and energy balances correlate the channels within a unit, a control loop ties a measurement to its setpoint, and a shared control lever drives its dependent channels together, so the few independent levers of A5 and the typed, correlated channels of A8 surface as blocks of the channel correlation graph. Agglomerative clustering on the train-normal correlation graph groups the channels into nested communities $G$ (dendrogram subtrees of size 3 to 25 channels) that serve as data-driven proxies for these subsystems, with no supervision and no plant schematic.

Faults inherit the same locality: a stuck valve, a drifting analyzer, or a fouled sensor perturbs its own subsystem while the rest of the plant runs normally, so an incipient fault is typically confined to one community, a handful of channels out of hundreds. A single global density dilutes such a local violation across the full dimensionality (a single-group shift is a vanishing fraction of WADI's 738-dimensional feature, 123 channels × 6 statistics), whereas a per-community density concentrates it. For each community $G$ we therefore fit the full detector of §4.1–4.3 on that community's channels alone and read its calibrated surprise $s_G$ (the train-normal upper-tail negative log-probability), and we include the global detector as an unfactorized *null* expert so the detector retains sensitivity to dense whole-system faults. Summing the per-community surprises gives a simple pooled score (because the communities are nested subtrees and can overlap, this sum is an over-complete pooling rather than an exact factorization),

$$-\log p(x) \approx \sum_{G \in \mathcal{C}} s_G\,(x),$$

and the detector scores a window by a weighted combination $s(x) = \bigoplus_{G \in \mathcal{C}} w_G\, s_G(x)$. The weight $w_G$ grows with the community's internal cohesion (its mean absolute pairwise channel correlation), because a density violation inside a tightly coupled subsystem (A8) is more diagnostic than the same deviation in a loosely coupled one. The combiner $\bigoplus$ must adapt to the fault's unknown extent: summation is optimal when the fault perturbs many communities, a maximum when it is confined to one, and Higher Criticism [34], a sparse-signal statistic that asks whether more communities are mildly surprised than chance allows, interpolates between the two without assuming the extent in advance. Writing the per-community upper-tail p-values for the $S$ community experts in ascending order $p_{(1)} \le \cdots \le p_{(S)}$, the statistic is

$$\mathrm{HC} = \max_{i: p_{(i)} < 1/2} \frac{\sqrt{S}\,(i/S - p_{(i)})}{\sqrt{p_{(i)}(1 - p_{(i)})}},$$

over the surprising tail. Cohesion weighting raises each p-value to the power $w_G/\bar{w}$ before sorting, with $w_G = \mathrm{coh}_G \sqrt{|\,G\,|}$ (community cohesion times root size) and $\bar{w}$ their mean, so a violation in a tight, large subsystem weighs more; every p-value is a train-normal upper-tail probability, so the combiner is calibrated on normal alone. The factorization also localizes: the per-community surprises rank the plant's subsystems by how implicated they are, turning a plant-wide alarm into a subsystem-level lead for the operator. The reported **LatAD (regime-community)** combines the communities by cohesion-weighted Higher Criticism and fuses that with the null expert by summing the two in train-normal z-score units; the single-latent **LatAD (global density)** of §4.1–4.3 is the degenerate case $\mathcal{C} = \{\text{all channels}\}$ and is reported as the base ablation. The detector occupies one point on each of four design axes: representation (VaDE), density estimation (itself two decisions, community construction and surprise aggregation), scoring heads, and calibration. Appendix A ablates the density-estimation axis.

## 5 Experimental methodology

### *5.1 Datasets*

We use three real CPS datasets and no synthetic data in the results tables.

- **WADI** [31], a water-distribution testbed, 123 sensor/actuator channels. Roughly a quarter of channels are near-constant in normal operation; these are attack *carriers*, not noise, and are kept. A few channels carry physically impossible standardized values (a channel constant at zero in the 14-day normal record, another recalibrated between files); standardized features are clipped to ±10σ to cap such sensor glitches. WADI is downsampled 10× on both streams, with the train and test horizons matched.
- **HAI** [33], a hardware-in-the-loop industrial control system testbed (HAI 20.07), 59 channels, a multi-process plant. No clipping (its large excursions are real attack signal).
- **SWaT** [32], the Secure Water Treatment testbed from iTrust, a 51-channel six-stage water-treatment plant with real executed attacks. No clipping. SWaT is downsampled 10× on both streams, with the train and test horizons matched. Because SWaT's attack record is a continuous attack period, the first 2% of the normal record (plant warm-up) is dropped and the last 20% of the normal record is held out to supply the normal windows of the test set; all calibration still uses train-normal only.

Temporal 10× downsampling of WADI and SWaT (both natively 1 Hz over many days) is standard practice on these benchmarks: it matches the 10-second median protocol of GDN [24] and lies within the downsampling range USAD [22] reports as leaving detection performance largely unchanged. HAI is used at native resolution.

### 5.2 *Windows and features*

Each raw channel is standardized to zero mean and unit variance using train-normal statistics only (A8). We then slide a fixed window of length $W = 60$ with stride 30 across the standardized stream. Each window is summarized by six per-channel statistics: mean (level), standard deviation (variability), minimum, maximum, first-to-last difference (net trend), and range, concatenated into one flat feature vector, so a $C$-channel window becomes a $6C$-dimensional input. A window is labeled anomalous if more than 5% of its timesteps are attack timesteps. Model inputs are additionally per-feature standardized on train-normal before the encoder. The window size and stride are unified across all three datasets. The reported model uses these static per-channel features; explicit multiscale temporal features (slope, velocity, spectral band power) motivated by A9 are left to future work (§7).

**Model configuration.** The VaDE uses $K$ mixture components and latent dimension set per dataset (WADI 20 and 10, HAI 40 and 16, SWaT 40 and 16); the density head is a diagonal mixture of 80 components (capped at one tenth of the train-window count), the residual head is reduced to 30 dimensions, and windows are length 60 with stride 30. The basin head uses $R = 16$ perturbations at noise scale 0.5σ, dead-zone $\delta = 0.15$ and $\lambda_0 = 2.5$; since the ambiguous-window fraction is ≈ 0.05 on all three datasets, below $\delta$, its scale is zero throughout. Correlation communities are agglomerative-clustering (average-linkage) subtrees of size 3 to 25 on the train-normal channel-correlation graph, yielding 45, 28, and 26 communities on WADI, HAI, and SWaT. All settings are shared across datasets except the per-dataset mixture size and latent dimension.

### 5.3 *Difficulty stratification*

A trivial detector scores each window by the maximum absolute standardized per-channel window mean, $\max_{ch} |z|$ (here $z$ is a standardized per-channel value, not the latent $z$ of §4.3), the simplest "is any channel out of its normal range" rule. An anomaly window is labeled **easy** if this trivial score exceeds the 99th percentile of train-normal, and **difficult** otherwise. Difficult windows are those this max-mean rule does not separate; §6 confirms the HAI advantage persists under a stronger split that also removes windows flagged by any of the six per-channel statistics, so difficulty is not defined by a single hand-picked statistic. We report all methods on three subsets: **Easy**, **Difficult**, and **All**. In every subset the positive class is that subset's anomaly windows and the negative class is the full set of normal windows, so the subsets differ only in which anomalies they include. The difficult column is the discriminative one.

### 5.4 *Comparable metrics*

We report two raw, point-wise metrics (each scored window is one evaluation point, with no point adjustment) and are explicit about their comparability across subsets.

- **AUROC** (the probability that a randomly chosen anomaly window outranks a randomly chosen normal window, 0.5 being chance) is prevalence-independent, hence directly comparable across Easy/Difficult/All, and requires no threshold. It is our primary cross-subset metric, reported as a five-seed mean for the learned detectors.
- **F1** is the best raw point-wise F1 over score thresholds (an oracle, field-standard threshold, disclosed plainly), computed per seed and averaged over the five seeds for the learned detectors. Because the subsets differ in positive count it is not directly comparable across them and serves as a secondary, within-subset figure.

All model fitting and all calibration (density head, nearest-component reference, residual precisions and gate, basin scale and reference, and every z-normalization) use train-normal data only. Test data never enters training or calibration. AUROC is leak-free; only the F1 threshold uses the test-swept oracle, which we flag.

### 5.5 *Baselines and SOTA re-computation*

Classical baselines are Isolation Forest [20] and a deep autoencoder, both on the same window features, together with a linear cross-channel baseline (LinRes): a leave-one-channel-out linear regression that predicts each channel's window summary from the others, with discrete actuator channels one-hot encoded, scored by its residual. For modern SOTA we re-run USAD [22] and TranAD [23] through the TranAD evaluation harness over five seeds and score them with the *same raw point-wise metrics*, aggregating their per-timestep scores onto our window grid. For HAI, the SOTA harness produces scores at a downsampled rate, so we map them to our full-resolution window grid by an integer-ratio correction; our detector and all local baselines use HAI at native resolution. We also re-run the graph-based GDN [24] under the identical protocol; its per-window graph training completed within our compute budget only on SWaT, where it matches the other deep detectors (difficult-subset AUROC 0.871, below ours). We do *not* apply point adjustment to any method. This is deliberate: point adjustment inflates F1 so severely that a random score outscores every deep model (§2), so a fair comparison must use raw metrics for all methods, including the published SOTA. We also report a diagnostic *trivial* baseline, the maximum absolute standardized per-channel window mean, the same score that defines the difficulty split (§5.3), whose behavior by construction is strong on Easy and near chance on Difficult.

## 6 Results

Table 3 reports AUROC and best raw point-wise F1 for every method on the three datasets and the three subsets. The headline model is **LatAD (regime-community)** (§4.4). Bold marks the best AUROC within a column across all methods; AUROC for the learned detectors is reported as mean±standard deviation over five seeds.

**Table 3.** Per-dataset, per-subset results [AUROC, F1] on three real CPS attack benchmarks (higher is better; AUROC is leak-free, F1 is the best raw point-wise F1 over thresholds). Learned detectors report AUROC as mean±standard deviation over five seeds (the standard deviation is omitted where it rounds to 0.000); GDN, whose per-window graph training completed only on SWaT within our compute budget, is single-run on SWaT. Bold marks the best AUROC in each dataset×subset column (ties bolded jointly). **LatAD (regime-community)** is the headline detector (§4.4); **LatAD (global density)** is its single-latent base ablation. Baselines are defined in §5.5.

| s | Group | All | | Easy | | Difficult | |
|---|---|---|---|---|---|---|---|
| | | AUROC | F1 | AUROC | F1 | AUROC | F1 |
| | | **WADI** (anom 56 = 37 easy + 19 difficult) | | | | | |
| trivial max\|z\| | Baseline | 0.558 | 0.000 | 0.687 | 0.000 | 0.307 | 0.000 |
| Isolation Forest | Baseline | 0.726±0.007 | 0.378 | 0.751±0.010 | 0.378 | 0.677±0.011 | 0.150 |
| AutoEncoder | Baseline | 0.743±0.001 | 0.611 | 0.907±0.002 | 0.724 | 0.425±0.005 | 0.223 |
| LinRes (one-hot) | Baseline | 0.595 | 0.214 | 0.699 | 0.220 | 0.392 | 0.182 |
| USAD | SOTA | 0.700±0.001 | 0.519 | 0.903±0.002 | 0.652 | 0.305±0.004 | 0.030 |
| TranAD | SOTA | 0.717±0.003 | 0.572 | 0.918±0.002 | 0.717 | 0.325±0.004 | 0.026 |
| LatAD (global density) | Ours | 0.792±0.012 | 0.565 | 0.845±0.007 | 0.680 | 0.690±0.027 | 0.222 |
| **LatAD (regime-community)** | Ours | 0.862±0.017 | 0.692 | 0.896±0.012 | 0.762 | 0.796±0.041 | 0.367 |
| | | **HAI** (anom 652 = 485 easy + 167 difficult) | | | | | |
| trivial max\|z\| | Baseline | 0.806 | 0.666 | 0.966 | 0.800 | 0.340 | 0.000 |
| Isolation Forest | Baseline | 0.844±0.011 | 0.418 | 0.919±0.016 | 0.468 | 0.627±0.005 | 0.059 |
| AutoEncoder | Baseline | 0.923±0.001 | 0.733 | 0.980 | 0.803 | 0.757±0.003 | 0.340 |
| LinRes (one-hot) | Baseline | 0.779 | 0.499 | 0.846 | 0.605 | 0.586 | 0.121 |
| USAD | SOTA | 0.843±0.005 | 0.691 | 0.969±0.001 | 0.823 | 0.477±0.018 | 0.018 |
| TranAD | SOTA | 0.834±0.002 | 0.692 | 0.968 | 0.824 | 0.444±0.010 | 0.015 |
| LatAD (global density) | Ours | 0.933±0.007 | 0.714 | 0.975±0.004 | 0.780 | 0.811±0.016 | 0.365 |
| **LatAD (regime-community)** | Ours | 0.949±0.004 | 0.756 | 0.983±0.002 | 0.822 | 0.849±0.018 | 0.364 |
| | | **SWaT** (anom 182 = 144 easy + 38 difficult) | | | | | |
| trivial max\|z\| | Baseline | 0.988 | 0.972 | 1.000 | 1.000 | 0.943 | 0.848 |
| Isolation Forest | Baseline | 0.959±0.003 | 0.914 | 0.987±0.002 | 0.960 | 0.853±0.011 | 0.659 |
| AutoEncoder | Baseline | 0.987±0.001 | 0.933 | 1.000 | 0.997 | 0.939±0.006 | 0.648 |

| s | Group | All | | Easy | | Difficult | |
|---|---|---|---|---|---|---|---|
| | | AUROC | F1 | AUROC | F1 | AUROC | F1 |
| LinRes (one-hot) | Baseline | 0.991 | 0.948 | 1.000 | 0.997 | 0.959 | 0.710 |
| USAD | SOTA | 0.972±0.001 | 0.948 | 1.000 | 0.997 | 0.867±0.005 | 0.718 |
| TranAD | SOTA | 0.972 | 0.950 | 1.000 | 0.997 | 0.867±0.001 | 0.737 |
| GDN | SOTA | 0.973 | 0.946 | 1.000 | 0.990 | 0.871 | 0.698 |
| LatAD (global density) | Ours | 0.991±0.002 | 0.950 | 0.999±0.002 | 0.974 | 0.960±0.006 | 0.761 |
| **LatAD (regime-community)** | Ours | 0.993±0.001 | 0.966 | 1.000 | 0.985 | 0.969±0.006 | 0.855 |

**Overall detection.** LatAD attains the best All-subset AUROC of any method on every dataset: WADI 0.862, HAI 0.949 and SWaT 0.993, clearing every baseline and both deep detectors on all three, including the strong classical baselines at the SWaT ceiling. The gain over the single-latent base (LatAD global density, 0.792 / 0.933 / 0.991) comes from factorizing the density over regime communities (§4.4), and is largest on the difficult subset, where it matters most.

**Difficult, joint-structure faults.** This is where the design earns its keep (Figure 1). The clearest demonstration is HAI: LatAD reaches 0.849 on the difficult subset (167 windows spanning 26 attack episodes), well above the AutoEncoder (0.757) and far above the deep detectors USAD (0.477) and TranAD (0.444), which sit near chance. The margin over the AutoEncoder is statistically significant: an episode-block bootstrap of the paired difference (resampling attack episodes and normal windows, averaged over five seeds) gives +0.092 with a 95% confidence interval of [0.046, 0.160] and P(difference ≤ 0) ≈ 0. On WADI the deep detectors collapse to 0.30–0.33 while LatAD reaches 0.796, a lead over Isolation Forest (0.677) that the same bootstrap places near significance (+0.12, 95% CI [−0.005, 0.257], P ≈ 0.03); the regime-community factorization is what turns the earlier single-latent tie (0.690) into this lead. On SWaT the difficult subset is only mildly difficult (the trivial rule already scores 0.943), so every method is strong (LatAD 0.969, the linear baseline 0.959) and its 38 windows come from a single attack episode, making SWaT a ceiling case rather than a discriminative test. The advantage over the deep detectors is largest exactly where the difficult anomalies are reconstructable yet improbable (WADI, HAI), and vanishes on SWaT, whose difficult anomalies are large multivariate deviations every method detects.

**Trivially separable anomalies.** On the easy subset every method is strong and the ranking is uninformative, by construction: the easy subset is exactly the anomalies a univariate max-mean rule already flags, and the trivial rule itself tops the easy column on SWaT (1.000).

**Validity of the difficulty split.** The trivial max|z| rule is strong on Easy by construction (SWaT 1.000, HAI 0.966, WADI 0.687) and drops to near chance on the difficult subset of the two benchmarks that carry genuine joint-structure faults (WADI 0.307, HAI 0.340): the difficult subset is precisely the anomalies this univariate max-mean rule cannot see, so a method that scores well there is detecting joint structure the rule misses. On SWaT the rule stays strong on the difficult subset (0.943), which is why SWaT is a strong overall benchmark but not a discriminative test of joint-structure detection.

**Robustness to a stronger difficulty definition.** The canonical split flags a window easy only when the maximum absolute standardized per-channel *mean* exceeds its 99th train-normal percentile, so a window could in principle be "difficult" yet be separable by a different univariate statistic. We therefore repeat the analysis with a stronger split that labels a window easy if the maximum over *all six* standardized per-channel statistics (mean, standard deviation, minimum, maximum, first-to-last difference, and range) exceeds its 99th train-normal percentile, removing any window a single univariate statistic separates. On the resulting HAI difficult subset (110 windows) the detector still leads every baseline (0.675, versus 0.622 for the linear cross-channel baseline, 0.604 for the AutoEncoder, and 0.528 for Isolation Forest), so its HAI advantage is not an artifact of the particular statistic that defines difficulty. WADI remains a tie (Isolation Forest 0.710, ours 0.697) and SWaT remains a ceiling case (trivial rule 0.929), consistent with the canonical split.

**The double-hard subset.** As a secondary robustness stress test we intersect the failures of the two simplest baseline families: a window is *double-hard* if it is an anomaly separated by *neither* the univariate max|z| rule *nor* the one-hot linear cross-channel predictor (LinRes), with both 99th-percentile thresholds fixed on train-normal only and

the detector under test playing no role in constructing the subset. Table 4 reports its per-method AUROC: LatAD attains the highest mean on all three datasets (WADI 0.728, HAI 0.819, SWaT 0.936). Statistical superiority is established on HAI, where 19 attack episodes support an episode-block bootstrap and the paired difference over the AutoEncoder is +0.089 (95% CI [0.020, 0.194]); the WADI subset rests on five episodes and is a numerical lead over Isolation Forest (0.728 vs 0.599, $P \approx 0.05$), and the 18 SWaT windows come from a single attack episode, so no episode-level generalization is claimed there. The subset removes ceiling signal rather than selecting a favorable slice: the trivial-rule AUROC falls from 0.943 on the difficult subset to 0.880 here, and the AutoEncoder collapses on WADI (0.336). By construction the two baselines that define the subset fail on it and are shown for completeness, not as an unbiased comparison.

**Table 4.** Double-hard subset AUROC (five-seed mean±standard deviation for the learned detectors). The double-hard subset keeps only anomaly windows separated by neither the univariate max|z| rule nor the linear cross-channel predictor, both thresholded on train-normal at the 99th percentile; it is a stress test independent of the detector under test (which plays no role in constructing it), not an unbiased comparison against those two filters. Windows / distinct attack episodes: WADI 11 / 5, HAI 84 / 19, SWaT 18 / 1. The detector has the highest mean AUROC on all three datasets; the difference is statistically supported only on HAI (episode-block bootstrap), a numerical lead on WADI (five episodes), and a single-episode result on SWaT.

| Method | WADI | HAI | SWaT |
|---|---|---|---|
| trivial max\|z\| | 0.283 | 0.349 | 0.880 |
| Isolation Forest | 0.599±0.013 | 0.635±0.012 | 0.789±0.011 |
| AutoEncoder | 0.336±0.007 | 0.730±0.003 | 0.890±0.010 |
| LinRes (one-hot) | 0.251 | 0.465 | 0.914 |
| USAD | 0.258±0.007 | 0.442±0.021 | 0.780±0.006 |
| TranAD | 0.272±0.002 | 0.418±0.009 | 0.777±0.003 |
| LatAD (global density) | 0.601±0.027 | 0.806±0.019 | 0.921±0.012 |
| **LatAD (regime-community)** | 0.728±0.077 | 0.819±0.020 | 0.936±0.013 |

**Deep SOTA under raw metrics.** Re-computed with raw point-wise metrics, the deep detectors USAD and TranAD collapse on the difficult subset of the two joint-structure benchmarks: over five seeds they fall to 0.477 and 0.444 on HAI and to 0.305 and 0.325 on WADI, near or below chance and far below our 0.796–0.849, even while they remain strong on Easy (0.90–0.97). This is the concrete face of the "illusion of success": models that look strong under point-adjusted F1 have little difficult-fault signal once the metric is raw.

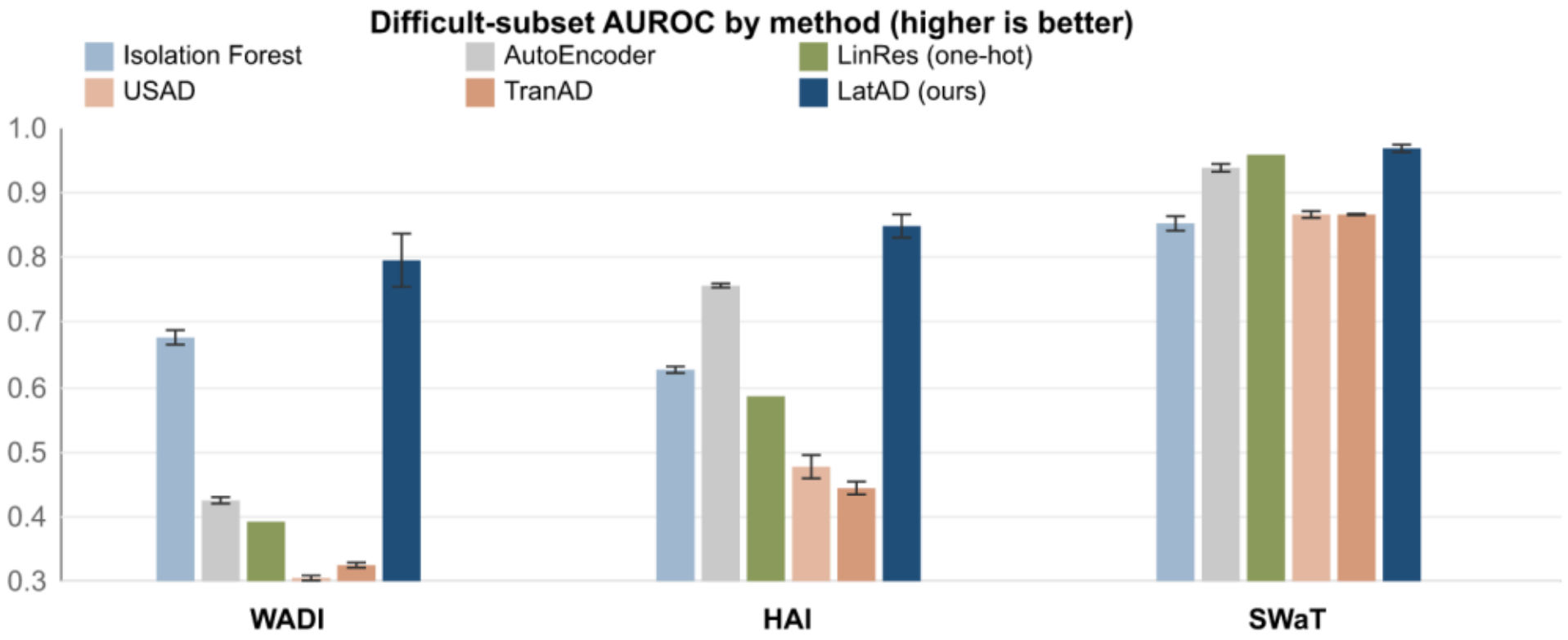

**Figure 1.** Difficult-subset AUROC by method (five-seed mean; whiskers show ±1 SD over seeds for the multi-seed detectors: Isolation Forest, AutoEncoder, USAD, TranAD, and LatAD). LatAD leads on all three datasets, significantly on HAI (0.849, a +0.09 margin over the AutoEncoder, 95% CI [0.046, 0.160]) and near-significantly on WADI (0.796 vs Isolation Forest 0.677); the deep detectors USAD and TranAD fall to 0.30–0.48 on WADI and HAI, whose difficult faults break the joint structure of normal operation, and rise only on SWaT, whose difficult faults are large deviations every method detects.

## 7 Discussion

**What the results show.** Under a fair protocol (raw point-wise metrics, difficulty stratification, and train-normal-only calibration) on three real CPS attack benchmarks, LatAD **attains the best overall AUROC on every dataset and leads the difficult subset of all three over five seeds**. The clearest demonstration is HAI, whose difficult faults break joint structure: LatAD reaches 0.849, a statistically significant margin over the next-best baseline (paired episode-bootstrap 95% CI [0.046, 0.160]), while the deep detectors USAD and TranAD fall near chance (0.44–0.48); on WADI the regime-community factorization turns a prior single-latent tie into a near-significant lead (0.796, $P \approx 0.03$), and SWaT is a ceiling. Modeling normal as a jointly learned latent plus explicit regime clustering, scored by probability and factorized over the plant's subsystems, is what delivers the advantage.

**Why reconstruction fails and density wins.** The deep detectors USAD and TranAD score by reconstruction, and reconstruction measures *reachability* (can the decoder reproduce this vector?), not *probability* (is this vector likely under normal?). On multimodal normal data the two diverge: a flexible decoder trained to reproduce several regimes interpolates between them, so an improbable between-regime or in-regime-tail combination sits inside the decoder's reachable range and reconstructs with low error. We verify this directly on WADI with a single-model diagnostic probe (one seed). The difficult anomalies stay on the normal manifold (PCA off-subspace residual AUROC 0.43) and reconstruct as well as normal (autoencoder reconstruction-error AUROC 0.43), yet the latent density flags them (0.74). A controlled probe makes the mechanism explicit: 1000 synthetic improbable combinations, each formed by averaging two normal windows from different regimes, reconstruct even better than real normal (reconstruction AUROC 0.08 against train-normal windows, WADI, one seed), while the latent density separates them cleanly (0.72). Scoring by density in the clustered latent replaces reachability with probability, which is exactly what reconstruction cannot see. This is why the base score omits reconstruction and scores in the jointly learned latent, with a high-resolution density head for non-Gaussian fringes (A4) and a rare-regime-safe nearest-component likelihood (A6); reconstruction is added back only through the optional, auto-gated residual head where it demonstrably helps (§4.3 iii).

**Why local density beats global density.** The reconstructable-but-improbable argument explains why we score density rather than reconstruction; a second mechanism explains why we score it per subsystem. A fault confined to one correlated channel group is, in the full-dimensional latent, a small perturbation of a mostly-normal state, so its surprise is averaged down by the many communities that stay normal, and the global density under-weights exactly the local violations that matter. Estimating density inside each community removes that dilution, and combining the per-community surprises by a sparsity-adaptive statistic recovers a signal carried by a few communities that a global sum would wash out. This is the mechanism behind the WADI gain (Table A1): the coordinated drift of a small analyzer group is improbable within its own subsystem yet unremarkable in the whole-plant density, so factorization turns the single-latent tie into a lead.

**Table 5.** Difficult-subset AUROC by score head, five-seed mean±standard deviation on all three datasets (an independent five-seed run). *recon* is the whitened reconstruction residual (the term the base score omits); *base* is density+nearest; the final row adds the auto-gated residual head, which is by definition the single-latent *LatAD (global density)* configuration and reproduces its Table 3 row within seed variation; its per-community application (§4.4) gives the headline detector. The basin head (iv) is inactive ($\lambda = 0$) on all three datasets and is omitted.

| Score head | WADI | HAI | SWaT |
|---|---|---|---|
| reconstruction residual (dropped term) | 0.475±0.004 | 0.695±0.004 | 0.965±0.002 |
| latent density | 0.694±0.027 | 0.802±0.011 | 0.952±0.008 |
| nearest-component NLL | 0.665±0.033 | 0.797±0.008 | 0.922±0.010 |
| base (density + nearest) | 0.685±0.030 | 0.801±0.010 | 0.947±0.007 |
| **base + resid (auto) = LatAD (global density)** | 0.685±0.030 | 0.820±0.010 | 0.960±0.002 |

Table 5 decomposes the score into its heads across five seeds and all three datasets. On WADI the reconstruction residual alone is near chance (0.475) while the latent density carries the signal (0.694); on HAI reconstruction is informative but the weakest head (0.695) and the density head dominates (0.802). The residual head, where the gate turns it on, lifts the base score exactly where reconstruction is informative (HAI 0.801→0.820, SWaT 0.947→0.960) and stays off on WADI. On SWaT reconstruction is in fact the strongest single head (0.965), consistent with SWaT's difficult faults being large deviations that reconstruction catches, which is why the gate keeps the residual head on there and why SWaT is a ceiling rather than a joint-structure test. The decomposition supports the mechanism at the head level: the latent density is what carries the joint-structure benchmarks, and reconstruction contributes only where it demonstrably generalizes.

**Auto-gating adapts one architecture to three datasets.** The winning signal is dataset-dependent, so rather than hand-select we gate two optional heads by purely train-normal signals: a responsibility-weighted whitened residual (gated by held-out-normal generalization; active on HAI and SWaT) and a basin-agreement head (gated by the ambiguous-normal ratio; a no-op on all three benchmarks here). The single global-density configuration then adapts itself: the residual head activates on HAI and SWaT, where the fault leaves a whitened-residual signal, and stays off on WADI, whose difficult faults are pure low-density pockets in the latent.

**The residual frontier on WADI.** A minority of WADI difficult windows are attack-onset/offset edges whose fault content is near-absent: 2–3$\sigma$ dips on a few correlated channels, sitting inside the normal tail, where normal CPS operation is itself noisy. None of the evaluated snapshot detectors, ours or the deep baselines, separates these at a low false-alarm budget; they are an intrinsic property of the data rather than of any one detector. Reaching them would require information beyond the instantaneous window, such as a physics-informed process residual, which we leave to future work.

**Trajectory assumptions A9–A10.** The reported model is window-only and leverages the instantaneous-structure assumptions A1–A8. Extending it with multiscale-temporal features (A9) and history-conditioned, path-dependent scoring (A10) is the natural next step; the public benchmarks used here are snapshot-detectable, so exercising the trajectory assumptions requires data that contains a history-dependent fault.

**Alternative realizations and validation.** The MIIM assumptions fix the structure to model rather than the estimator, so the detector of §4 is one point in a larger design space. The regime-community realization (§4.4) exercises the factorization and sparsity-adaptive-combination routes; further routes remain open within the same assumption set: non-parametric or normalizing-flow densities per community, alternative unsupervised partitioners, and a deeper multiscale community hierarchy than the current nested-subtree construction. Because the community construction and the cohesion-weighted Higher Criticism combiner were developed on these three benchmarks, the regime-community realization should be validated on additional, untouched CPS systems before its per-dataset gains are treated as established; its build and calibration use train-normal data only, so the extension is directly testable on additional CPS systems.

**Limitations.** Three boundary conditions apply: three datasets, one trained configuration per dataset (evaluated over five seeds), and a single instantaneous representation. AUROC is threshold-free and reported as a five-seed mean; the F1 column is a best-threshold raw F1. The easy/difficult split is a property of one cut, the trivial detector's 99th-percentile-of-train-normal threshold; a different percentile would redraw the boundary between the subsets, though the gap between the trivial rule's Easy and Difficult performance is large enough that the split is not knife-edge. We report only raw point-wise metrics, which makes our numbers lower than point-adjusted leaderboard figures by design.

## 8 Conclusion

We restated the structural assumptions of CPS normal data as an explicit MIIM list (A1–A10) and built LatAD, a detector that models normal as a jointly learned latent plus explicit regime clustering, scores anomalies by probability in that latent rather than by reconstruction, and factorizes the density over the plant's correlation-community subsystems. Under a deliberately fair, raw-metric, difficulty-stratified protocol on three real CPS attack benchmarks, LatAD attains the best overall AUROC on every dataset and leads the difficult subset of all three, significantly on HAI (paired 95% CI [0.046, 0.160]) and near-significantly on WADI, where the reconstruction-based deep detectors collapse.

The mechanism is simple to state. Reconstruction scores reachability; a density in a clustered latent scores probability; and the CPS faults that matter are reconstructable but improbable. Treating multimodal normal as a single blob, or as a reconstruction target, is a modeling failure.

Two properties make the detector well-suited to condition monitoring and predictive maintenance: scoring against the nearest regime keeps rare-but-valid states (startups, rare manoeuvres) from paging an operator, and the regime-community density catches in-envelope, subsystem-local faults that stay inside every alarm limit. Extending LatAD to the trajectory assumptions (A9, A10) and validating the regime-community realization on additional CPS systems are the natural next steps.

**Author Contributions:** Conceptualization, methodology, software, validation, formal analysis, investigation, writing (original draft preparation), and writing (review and editing), A.A. and Y.A. The authors contributed equally to all aspects of the work. All authors have read and agreed to the published version of the manuscript.

**Data Availability Statement:** This study uses only publicly available community benchmark datasets and created no new data. WADI and SWaT are available from iTrust, Singapore University of Technology and Design (https://itrust.sutd.edu.sg/, access granted on request); HAI is available from its public release by ETRI. The trained model checkpoints (one per dataset) and the source code supporting the reported results are openly archived on Zenodo at https://doi.org/10.5281/zenodo.21821524. Each checkpoint contains learned parameters only, not the underlying data; each is trained with a fixed seed and reproduces that seed's scores, and the released scripts reproduce the five-seed aggregates reported in the tables.

## Appendix A. Ablation over the design axes

LatAD is one point on each design axis of §4: representation, density estimation (single global versus a regime-community ensemble, itself construction plus aggregation), scoring heads, and calibration. Table A1 varies the density-estimation axes on the difficult subset, holding the rest fixed. Factorizing the global density over regime communities lifts all three datasets, most on WADI, whose sparse coordinated-drift faults a global density dilutes; fitting the full per-community scoring heads (adding the A8 whitened residual inside each subsystem) further lifts HAI, where the hard valve and flow deviations are correlation-conditional; and among the sparsity-adaptive combiners, cohesion-weighted Higher Criticism is the only one that leads on all three, because it weights a violation by the coupling strength of the subsystem it occurs in (A8) while adapting to the fault's extent.

**Table A1.** Design-axis ablation, difficult-subset AUROC (five-seed mean). Top block: the factorization axis (single global density, then the regime-community ensemble with density-only versus full per-community heads). Bottom block: the aggregation axis over the full-stack experts (how per-community surprises are combined). Every aggregation-block row is fused with the global null expert: the first four sum the community combiner and the null expert in train-normal z-score units (the headline fusion), while the last fuses unweighted Higher Criticism with the null expert by a maximum instead. The headline detector is the full per-community, cohesion-weighted Higher Criticism row.

| Ablation | WADI | HAI | SWaT |
|---|---|---|---|
| **Factorization (density-estimation axis)** | | | |
| LatAD (global density), no factorization | 0.690 | 0.811 | 0.960 |
| LatAD (regime-community), density-only experts | 0.799 | 0.814 | 0.968 |
| **LatAD (regime-community), full per-community heads** | 0.796 | 0.849 | 0.969 |
| **Aggregation (over full-stack experts)** | | | |
| sum (of community surprises) | 0.747 | 0.828 | 0.968 |
| max (most-surprised community) | 0.749 | 0.802 | 0.964 |
| Higher Criticism (unweighted) | 0.753 | 0.828 | 0.968 |
| **cohesion-weighted Higher Criticism (headline)** | 0.796 | 0.849 | 0.969 |
| Higher Criticism with null expert, max-fused | 0.744 | 0.846 | 0.964 |

## References

1. Chandola, V.; Banerjee, A.; Kumar, V. Anomaly detection: A survey. *ACM Comput. Surv.* **2009**, *41*, 15. https://doi.org/10.1145/1541880.1541882.
2. Ruff, L.; Kauffmann, J.R.; Vandermeulen, R.A.; Montavon, G.; Samek, W.; Kloft, M.; Dietterich, T.G.; Müller, K.-R. A unifying review of deep and shallow anomaly detection. *Proc. IEEE* **2021**, *109*, 756–795. https://doi.org/10.1109/JPROC.2021.3052449.
3. Kim, S.; Choi, K.; Choi, H.-S.; Lee, B.; Yoon, S. Towards a rigorous evaluation of time-series anomaly detection. In Proceedings of the AAAI Conference on Artificial Intelligence, Virtual, 22 February–1 March 2022; Volume 36, pp. 7194–7201. https://doi.org/10.1609/aaai.v36i7.20680.

4. Garg, A.; Zhang, W.; Samaran, J.; Savitha, R.; Foo, C.-S. An evaluation of anomaly detection and diagnosis in multivariate time series. *IEEE Trans. Neural Netw. Learn. Syst.* **2022**, *33*, 2508–2517. https://doi.org/10.1109/TNNLS.2021.3105827.
5. Wu, R.; Keogh, E.J. Current time series anomaly detection benchmarks are flawed and are creating the illusion of progress. *IEEE Trans. Knowl. Data Eng.* **2023**, *35*, 2421–2429. https://doi.org/10.1109/TKDE.2021.3112126.
6. Pang, G.; Shen, C.; Cao, L.; van den Hengel, A. Deep learning for anomaly detection: A review. *ACM Comput. Surv.* **2021**, *54*, 38. https://doi.org/10.1145/3439950.
7. Blázquez-García, A.; Conde, A.; Mori, U.; Lozano, J.A. A review on outlier/anomaly detection in time series data. *ACM Comput. Surv.* **2021**, *54*, 56. https://doi.org/10.1145/3444690.
8. Hu, Y.; Yang, A.; Li, H.; Sun, Y.; Sun, L. A survey of intrusion detection on industrial control systems. *Int. J. Distrib. Sens. Netw.* **2018**, *14*, 1550147718794615. https://doi.org/10.1177/1550147718794615.
9. Doshi, K.; Abudalou, S.; Yilmaz, Y. Reward once, penalize once: Rectifying time series anomaly detection. In Proceedings of the International Joint Conference on Neural Networks (IJCNN), Padua, Italy, 18–23 July 2022. arXiv:2203.05167.
10. Ghorbani, R.; Reinders, M.J.T.; Tax, D.M.J. PATE: Proximity-aware time series anomaly evaluation. In Proceedings of the 30th ACM SIGKDD Conference on Knowledge Discovery and Data Mining (KDD), Barcelona, Spain, 25–29 August 2024; pp. 872–883. https://doi.org/10.1145/3637528.3671971.
11. Pinet, M.; Cumin, J.; Berlemont, S.; Vaufreydaz, D. Anomalies in multivariate time series benchmarks are mostly univariate. *arXiv* **2026**, arXiv:2606.02670. https://arxiv.org/abs/2606.02670.
12. Qiu, X.; Li, Z.; Wu, X.; Hu, J.; Guo, C.; Yang, B.; Jensen, C.S. TAB: Unified benchmarking of time series anomaly detection methods. *Proc. VLDB Endow.* **2025**. arXiv:2506.18046.
13. Bouman, R.; Heskes, T. Autoencoders for anomaly detection are unreliable. *arXiv* **2025**, arXiv:2501.13864. https://arxiv.org/abs/2501.13864.
14. Gong, D.; Liu, L.; Le, V.; Saha, B.; Mansour, M.R.; Venkatesh, S.; van den Hengel, A. Memorizing normality to detect anomaly: Memory-augmented deep autoencoder for unsupervised anomaly detection. In Proceedings of the IEEE/CVF International Conference on Computer Vision (ICCV), Seoul, Republic of Korea, 27 October–2 November 2019; pp. 1705–1714. arXiv:1904.02639.
15. Sarfraz, M.S.; Chen, M.-Y.; Layer, L.; Peng, K.; Koulakis, M. Position: Quo vadis, unsupervised time series anomaly detection? In Proceedings of the 41st International Conference on Machine Learning (ICML), Vienna, Austria, 21–27 July 2024. arXiv:2405.02678.
16. Wagner, D.; Michels, T.; Schulz, F.C.F.; Nair, A.; Rudolph, M.; Kloft, M. TimeSeAD: Benchmarking deep multivariate time-series anomaly detection. *Trans. Mach. Learn. Res.* **2023**. https://openreview.net/forum?id=iMmsCI0JsS.
17. Jiang, Z.; Zheng, Y.; Tan, H.; Tang, B.; Zhou, H. Variational deep embedding: An unsupervised and generative approach to clustering. In Proceedings of the 26th International Joint Conference on Artificial Intelligence (IJCAI), Melbourne, Australia, 19–25 August 2017; pp. 1965–1972. https://doi.org/10.24963/ijcai.2017/273.
18. Zong, B.; Song, Q.; Min, M.R.; Cheng, W.; Lumezanu, C.; Cho, D.; Chen, H. Deep autoencoding Gaussian mixture model for unsupervised anomaly detection. In Proceedings of the International Conference on Learning Representations (ICLR), Vancouver, BC, Canada, 30 April–3 May 2018. https://openreview.net/forum?id=BJJLHbb0-.
19. Ledoit, O.; Wolf, M. A well-conditioned estimator for large-dimensional covariance matrices. *J. Multivar. Anal.* **2004**, *88*, 365–411. https://doi.org/10.1016/S0047-259X(03)00096-4.
20. Liu, F.T.; Ting, K.M.; Zhou, Z.-H. Isolation forest. In Proceedings of the 2008 Eighth IEEE International Conference on Data Mining (ICDM), Pisa, Italy, 15–19 December 2008; pp. 413–422. https://doi.org/10.1109/ICDM.2008.17.
21. Sakurada, M.; Yairi, T. Anomaly detection using autoencoders with nonlinear dimensionality reduction. In Proceedings of the MLSDA 2014 2nd Workshop on Machine Learning for Sensory Data Analysis, Gold Coast, Australia, 2 December 2014; pp. 4–11. https://doi.org/10.1145/2689746.2689747.
22. Audibert, J.; Michiardi, P.; Guyard, F.; Marti, S.; Zuluaga, M.A. USAD: Unsupervised anomaly detection on multivariate time series. In Proceedings of the 26th ACM SIGKDD International Conference on Knowledge Discovery & Data Mining (KDD), Virtual Event, CA, USA, 23–27 August 2020; pp. 3395–3404. https://doi.org/10.1145/3394486.3403392.
23. Tuli, S.; Casale, G.; Jennings, N.R. TranAD: Deep transformer networks for anomaly detection in multivariate time series data. *Proc. VLDB Endow.* **2022**, *15*, 1201–1214. https://doi.org/10.14778/3514061.3514067.
24. Deng, A.; Hooi, B. Graph neural network-based anomaly detection in multivariate time series. In Proceedings of the AAAI Conference on Artificial Intelligence, Virtual, 2–9 February 2021; Volume 35, pp. 4027–4035. https://doi.org/10.1609/aaai.v35i5.16523.

25. Feng, Y.; Zhang, W.; Fu, Y.; Jiang, W.; Zhu, J.; Ren, W. SensitiveHUE: Multivariate time series anomaly detection by enhancing the sensitivity to normal patterns. In Proceedings of the 30th ACM SIGKDD Conference on Knowledge Discovery and Data Mining (KDD), Barcelona, Spain, 25–29 August 2024; pp. 782–793. https://doi.org/10.1145/3637528.3671919.
26. Wu, X.; Qiu, X.; Li, Z.; Hu, Y.; Guo, C.; Xu, X.; Yang, B. CATCH: Channel-aware multivariate time series anomaly detection via frequency patching. In Proceedings of the International Conference on Learning Representations (ICLR), Singapore, 24–28 April 2025. arXiv:2410.12261.
27. Gildish, E.; Grebshtein, M.; Aperstein, Y.; Makienko, I. Vibration-based estimation of gearbox operating conditions: Machine learning approach. In Proceedings of the 2023 International Conference on Control, Automation and Diagnosis (ICCAD), Rome, Italy, 10–12 May 2023; pp. 1–6. https://doi.org/10.1109/ICCAD57653.2023.10152303.
28. Gildish, E.; Grebshtein, M.; Aperstein, Y.; Makienko, I. Vibration signal decomposition using dilated CNN. *Proc. Annu. Conf. PHM Soc.* **2023**, *15*. https://doi.org/10.36001/phmconf.2023.v15i1.3502.
29. Gildish, E.; Grebshtein, M.; Aperstein, Y.; Kushnirski, A.; Makienko, I. Helicopter bolt loosening monitoring using vibrations and machine learning. *PHM Soc. Eur. Conf.* **2022**, *7*, 146–155. https://doi.org/10.36001/phme.2022.v7i1.3322.
30. Shavit, D.; Davidovits, M.; Kushnirsky, A.; Aperstein, Y. Temporal causality-based feature selection for fault prediction in rotorcraft flight controls. *IFAC-PapersOnLine* **2022**, *55*, 235–239. https://doi.org/10.1016/j.ifacol.2022.04.199.
31. Ahmed, C.M.; Palleti, V.R.; Mathur, A.P. WADI: A water distribution testbed for research in the design of secure cyber physical systems. In Proceedings of the 3rd International Workshop on Cyber-Physical Systems for Smart Water Networks (CySWATER), Pittsburgh, PA, USA, 21 April 2017; pp. 25–28. https://doi.org/10.1145/3055366.3055375.
32. Mathur, A.P.; Tippenhauer, N.O. SWaT: A water treatment testbed for research and training on ICS security. In Proceedings of the 2016 International Workshop on Cyber-Physical Systems for Smart Water Networks (CySWater), Vienna, Austria, 11 April 2016; pp. 31–36. https://doi.org/10.1109/CySWater.2016.7469060.
33. Shin, H.-K.; Lee, W.; Yun, J.-H.; Kim, H. HAI 1.0: HIL-based augmented ICS security dataset. In Proceedings of the 13th USENIX Workshop on Cyber Security Experimentation and Test (CSET), Virtual, 10 August 2020. https://www.usenix.org/conference/cset20/presentation/shin.
34. Donoho, D.; Jin, J. Higher criticism for detecting sparse heterogeneous mixtures. *Ann. Stat.* **2004**, *32*, 962–994. https://doi.org/10.1214/009053604000000265.